# The Self-Organizing Symbiotic Agent

Babak Hodjat            Makoto Amamiya

Department of Intelligent Systems
Graduate School of Information Science and Electrical Engineering
Kyushu University
6-1 Kasugakoen, Kasuga-shi
Fukuoka 816, Japan
http://www_al.is.kyushu-u.ac.jp/~bobby/index.html
e-mail: Bobby@diana.is.kyushu-u.ac.jp

**Abstract.**  In [2] a general framework for the study of Emergence and hyper-structure was presented. This approach is mostly concerned with the description of such systems. In this paper we will try to bring forth a different aspect of this model we feel will be useful in the engineering of agent based solutions, namely the symbiotic approach. In this approach a self-organizing method of dividing the more complex "main-problem" to a hyper-structure of "sub-problems" with the aim of reducing complexity is desired. A description of the general problem will be given along with some instances of related work. This paper is intended to serve as an introductory challenge for general solutions to the described problem.

**Topics:** Agents Theories, Agent Architectures.

# 1. Introduction

> *"A fundamental concept of evolution is the belief in the natural progression from the simple, to the more complex"* [5]

In Reinforcement Learning [15] an agent, or a population of agents interact with their environment (be it a common environment or one dedicated to an individual agent). Using trial and error interactions the agents are given the task of maximizing the sum of the reinforcements received when starting from some initial state and proceeding to a terminal state. It is assumed that there exists a mapping from state/action pairs to reinforcements, which is given to an agent as the result of its performing an action in a given state (reward, in the form of a scalar value). Therefore the "goal "of the reinforcement learning system is defined using the concept of a reinforcement function, which is the exact function of future reinforcements the agent seeks to maximize.

Many different Reinforcement Learning systems have been studied and some real world problems have been solved using this methodology [15]. One of the main problems encountered is the seemingly exponential reduction in the performance of the agents as the number of tasks and/or states increase. All currently published non-parametric learning algorithms utilize search strategies that are impractical in high dimensional space [14]. This is mainly due to the fact that the complexity of the search space has increased to a degree that the agent is unable to find maximizing functions suitable for the rapid approach to the desired solution. It seems that making agents that can learn and solve small-scale problems is much easier than upgrading this method to deal with problems of higher complexity.

In the symbiotic approach we want to find an automatic[1] way of dividing the more complex "main-problem" to a hyper-structure [2] of "sub-problems" such that the lowest level of the structure be satisfactorily learnable by agents. In other words, instead of maintaining a population of full solutions to the problem, in the symbiotic approach each population member is only a *partial solution*; The complete solution emerging from the interactions of the different agents. Partial solutions can be characterized as *specializations*. Instead of solving the entire problem, partial solutions specialize towards one aspect of the problem. Thus, many different areas of the solution space can be searched concurrently, allowing faster solutions to harder problems [29].

This paper attempts to introduce and describe this approach. We will first present a mathematical model for the hyper-structure and its expected emergent qualities deriving from previous work by [2]. Certain aspects that will have to be taken into account will also be presented. The advantages and the whole point of this method will then be discussed. Related work done in this domain seem to be very limited. We think this is partly due to the fact that general guidelines for this approach and their benefits have not been explained before. Finally some areas worth exploring for other possible solutions to this problem are noted.

# 2. Hyper-structures and Emergence

> *"One can think of organizing populations of machines or robots in such a way that their interactive dynamics lead to completely new properties and behavior."* [3]

If a property can be observed in the dynamics, but not at the level of the fundamental first order interacting structures, it is defined to be *emergent*. A third order structure is defined through the interaction of second order structures forming a new observable, not found at the lower levels. It is well known that second order structures occur relatively easy in simulation so the problem is how to proceed to third and higher orders without external interference [4]. "Whether more object complexity always will be necessary to obtain yet higher order emergent structures - or whether there is an object complexity limit above which any emergent level can be produced, we do not know. We would like to believe that the latter is true" [32].

---

[1] By automatic we are implying self-organization.

[2] defines the $N^{th}$ order structure, which in principle can depend on all lower level structure families in a cumulative way, as

$$S_N = R(S_{i_{N-1}}^{N-1}, Obs^{N-1}, Int^{N-1}), i_{N-1} \in J_{N-1}$$

where R is the result of a construction process on lower order structures with *Obs* as their observational mechanism (to obtain the properties of the structures), and *Int* representing a family of interactions using the properties registered under the observation (which could be a dynamic process). *i* belongs to some index set (finite or infinite) namely *J* and is an index to a family of structures[2] of a certain order (A complete explanation of this definition and some examples can be found in [2]).

Baas calls this whole process "the higher-order structure principle" the result of which he calls a *hyper-structure* – in this case of order *N*. We shall define the function *SO* as a function that will return the numeric value of the order of a structure (i.e., $SO(S^N)=N$).

Through this level structure we can distinguish and classify orders of emergence and complexity. *P* is an emergent property of $S^N$ iff

$$(P \in Obs^N(S^N)) \wedge (\forall i_{N-1} \mid P \notin Obs^N(S_{i_{N-1}}^{N-1}))$$

It must be noted that overlapping aggregates are allowed, and also that the creation of high-level interactions may cause changes in lower-level interaction *(Fig 1)*.

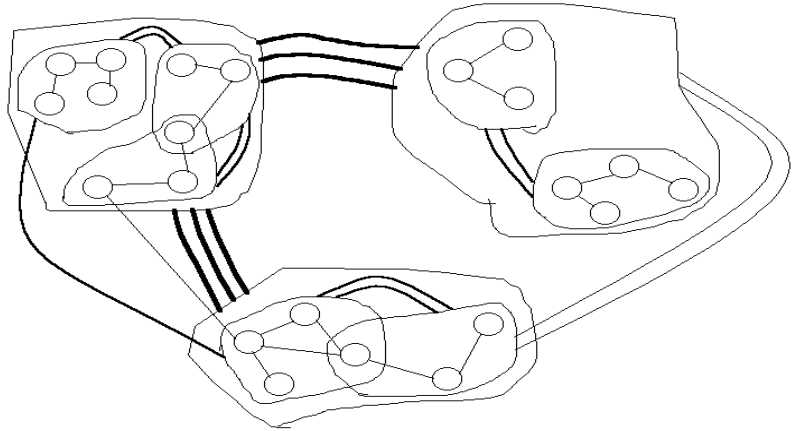

**Fig.1.** Hyper-structure: Example of third-order interaction graph of a hyper-structure allowing cumulative interactions and even overlapping aggregates [2].

As an extension of [34]'s thoughts about complexity and hierarchies, [2] is convicted that "complexity often takes the form of hyper-structure". In other words, for every level, the information-carrying capacity increases.

It seems reasonable to require that a self-organizing system should exhibit emergent properties. Baas names a twofold point of view with regard to models where higher-order hyper-structures naturally emerge: To understand the *generation* of hyper-structure from first principles, and to understand how to guide (externally) the *design* of the hyper-structure (man-made evolution [3]). We will be introducing another aspect of this second view.

---

[2] The primitives or first order structures are $S_{i_1}^1$, $i_1 \in J_1$.

## 3. A general definition of the agents

Let us say we have a population of dynamic problem-solving agents, which are taken to be sufficiently efficient to solve problems of a certain complexity. We take $SS^r$ (Solution Structure of order $r$) to represent this structure with $PC(SS^r)$ as its potential complexity. A given problem is of an unknown order of complexity ($x$) higher than $PC(SS^r)$. We take this complexity to be $C(PS^x)$ (The complexity of order $x$ of the Problem Structure). The following is desired: The design of a population of agents that are able to self-organize into a hyper-structure $SS^N$ such that

$$PC(SS^N) \geq C(PS^x).$$

Deriving from [2], we will now try to set down a model of such a system.

To show the possibility of interaction between two structures $S_1$ and $S_2$ we will use the following operator:

$$S_1 \oplus S_2 \quad (S_1 \text{ can and may interact with } S_2 \text{ or } INT(S_1, S_2) ).$$

Some of the properties of this operator are as follows:

- $S_1 \oplus S_1$,
- $(S_1 \oplus S_2) \Rightarrow (S_2 \oplus S_1)$,
- $(S_1 \oplus S_2) \wedge (S_1 \oplus S_3) \Rightarrow (S_1 \oplus S_3)$.

We also need to define an operator to show the dependency between two structures. This concept will be shown as:

$$S_1 \leftarrow S_2 \quad (\text{The process or output of } S_2 \text{ is part of } S_1)$$

Some of the properties of this operator are as follows:

- $(S_1 \leftarrow S_2) \Rightarrow S_1 \oplus S_2$,
- $(S_1 \leftarrow S_2) \Rightarrow SO(S_1) > SO(S_2)$,
- $(S_1 \leftarrow S_2) \wedge (S_2 \leftarrow S_3) \Rightarrow (S_1 \leftarrow S_3)$.

According to the definition of hyper-structures, we can derive the following:

**Theorem 1**:

$$(S_1 \leftarrow S_2) \wedge (\forall S_1, \nexists X \mid (S_1 \leftarrow X) \wedge (SO(X) > SO(S_2))) \Rightarrow SO(S_1) = SO(S_2) + 1.$$

The proof of this theorem is not far fetched considering the fact that according to the definition of the dependency operator ($\leftarrow$),

$$(S_1 \leftarrow S_2) \Rightarrow SO(S_1) > SO(S_2) \Rightarrow SO(S_1) \neq SO(S_2)$$

Let us presume:

$$SO(S_1) = SO(S_2) + c \qquad \text{where } c > 1.$$

In this case, according to the definition of hyper-structures, there must exist a structure, $Y$, with:

$$SO(S_1) > SO(Y) > SO(S_2),$$

Such that:

$$S_1 \leftarrow Y$$

Which is impossible according to the givens of the theorem. Therefore we conclude that

$$c = 1.$$

At this point we will set out to define our model using the concepts introduced above. Let us say we have a population of homogenous interacting solution hyper-structures (*SS*) of order *r* collaboratively striving to solve a problem:

$$SS^r_1 \oplus SS^r_2 \oplus SS^r_3 \oplus ... \oplus SS^r_p, \quad \text{(where } p \leq \text{population limit)}$$

The nature of these interactions may be evolutionary (hence the limit on the population), communicational, merely an indirect effect through the problem space or common environment, or any combination of these forms. We will represent this population with *Pop(SS$^r$)*

The solution hyper-structures should have the ability to increase their order (*break*). To model this ability we will define *B(Pop(SS$^r$))*, which represents the moment at least one of the members of *Pop(SS$^r$)* has had an increase in its order. According to theorem 1, it is enough that at least two of the solution hyper-structures in *Pop(SS$^r$)* change their nature of interaction from $\oplus$ to $\leftarrow$ for *B(Pop(SS$^r$))* to have happened. So:

$$B(Pop(SS^r)) \Rightarrow Pop^2(SS^r) \quad \text{if} \quad \exists X, Y \in Pop(SS^r) \,/\, X \leftarrow Y$$

*Pop$^2$(SS$^r$)* is therefore a non-homogenous population of solution structures formed out of *Pop(SS$^r$)*, comprised of solution structures of orders *r* and *r* + 1:

$$SO(Pop^2(SS^r)) = SO(Pop(SS^r)) + 1$$

And accordingly:

$$SO(Pop^{n+1}(SS^r)) = SO(Pop^n(SS^r)) + 1$$

Similarly, to be able to use the concept of *B* for higher order populations we need to refine the definition:

$$B(Pop^2(SS^r)) \Rightarrow Pop^3(SS^r) \text{ if } \exists X, Y \in Pop^2(SS^r) \,/\, (SO(X)=SO(Y)=2) \wedge (X \leftarrow Y)$$

And accordingly:

$$B(Pop^n(SS^r)) \Rightarrow Pop^{n+1}(SS^r) \text{ if } \exists X, Y \in Pop^n(SS^r)/(SO(X)=SO(Y)=n) \wedge (X \leftarrow Y)$$

To use [2]'s notation, *B* could be defined as follows:

$$B(Pop^n(SS^r)) \Rightarrow Pop^{n+1}(SS^r) \text{ if } \quad \exists X, Y \in Pop^n(SS^r) \,/$$
$$((X \leftarrow Y) \in Obs^n(Pop^n(SSr))) \wedge ((X \leftarrow Y) \notin Obs^n(Pop^{n-1}(SS^r)))$$

This form clearly shows the emergent property of *B*.

Note that *B* is bounded by the *population limit*. Other boundaries could also be set for *B*'s application. With an effective breaking policy, the initial population of order *r* individuals will increase their order:

*Step 1: $Pop^1(SS^r)$,*
*Step 2: $B(Pop^1(SS^r)) \Rightarrow Pop^2(SS^r)$,*
*Step 3: $B(Pop^2(SS^r)) \Rightarrow Pop^3(SS^r)$,*
.
.
.
*Step n: $B(Pop^{n-1}(SS^r)) \Rightarrow Pop^n(SS^r)$,*

It must be noted here that the breaking process, being a property of the individuals in the population, need not necessarily occur with a constant frequency. This process can be directed to happen when the need for it is detected by the failure of the existing population to approach the solution of the problem with the desired speed. Another aspect worth noting is that the Breaking function should also have a reverse equivalent so as to dynamically reverse the breaking process and reduce the order of the solution hyper-structure when such an order is no longer needed. For example an evolutionary fine-tuning safeguard may be effective in this regard.

From now on we shall refer to these agents as <u>*S*</u>elf-<u>*O*</u>rganizing <u>*S*</u>ymbiotic <u>*Age*</u>nts (or *SOSAGEs*). SOSAGEs will therefore solve a given problem by breaking it down to smaller, more manageable pieces. Each SOSAGE or group of SOSAGEs will become specialized in managing parts of the interactions in this hyper-structure, the result of which would be the solution.

An example would be the behavior of the nervous system. Its essence lies in the envisaging of mature capabilities as the cumulative outcome of a large number of small advances in capability. Each such advance is assumed to be associated with a particular generative mechanism, so that the nervous system as a whole can be viewed as an organized collection of such generative mechanisms, each specialized to some particular generative process. Just as in the case of a factory, the products of one kind of generative mechanism are handed on to other such mechanisms for further constructions to be carried out [21].

As another example let us consider a Reinforcement Learning [15] agent with a set of input senses which it uses to choose from a menu of actions and thus solve a problem (e.g., box pushing). The standard approach (e.g., Q-Learning) would be for an agent or a population of agents, to learn to map sequences of senses to certain sequences of actions in order to maximize the sum of the reinforcements received from the environment (or solution evaluator). More specialized but practicable solutions for more complex problems break down the problem into a set of predefined behaviors, lowering the potential complexity required from the agents (e.g., Behavior-based Q-learning [24]). SOSAGEs are expected to achieve this complexity reduction automatically, hence automatically breaking the problem into behaviors and sub-behaviors until they are manageable by the learning agents in an efficient manner. In other words, individuals will group together in one population that can be in itself regarded as a higher-level individual.

Is a SOSAGE possible? We will show later in this paper that work has been done in this regard and some general solutions have been proposed that come close to achieving a SOSAGE. In [8] the authors give a mathematical model of cooperative hierarchies and prove that the problem of determining whether cooperation structures are available to achieve an agent's goal is NP-complete. What are the distinguishing features of problems that can be solved using SOSAGEs? This question remains to be dealt with.

# 4. Advantages of this approach

The SOSAGE approach is, by nature, a general solution method. Self-organization implies learning and the flexibility that is expected from a SOSAGE forces it to be problem independent. But the most important feature of this approach is its *scalability*. That is to say machine learning solutions that perform reasonably well in lower complexities can be applied to problems with higher magnitudes of complexity, and still be expected to converge in a relatively reasonable manner. It has been shown that in nature, biological life [20] breaks down complex problems in such a way, hence the advent of artificial evolutionary optimization may be justifiable.

For instance, in the Reinforcement Learning example, in order to cope with the scale up problem procedural representations of policies may be adopted automatically [17]. Procedural representations have two main advantages. First, they are implicit, allowing for good inductive generalization over a very large set of input states. Second, they facilitate *modularization*. Modularization promotes diversity and prevents the population from converging to a single solution, which in turn promotes *generalization* and *flexibility* [11]. Research has shown that methods that decompose a problem space improve learning performance in Reinforcement learning [7] [35].

Concrete proof of the advantages of this approach can only be given based on practical experiments, evidence of which will hopefully be presented in later papers.

# 5. Related work

We will consider work related to the theme of this paper in three main areas, Distributed Artificial Intelligence (DAI), Evolutionary Neural Networks, and Behavior Based Reinforcement Learning.

Theoretical work in DAI regarding cooperating experts has been in two main directions, namely *task sharing* and *result sharing* [37]. In task sharing, agents break the problem and request assistance from other agents in solving them. This is done when a task is too large for the agent to handle, or it is a task for which the agent has no expertise. Two approaches are usually considered in this regard. Either for the agents to describe the task to other agents known to be able to assist, or to describe them to all other agents (blackboard systems). In result sharing systems, expert agents periodically report partial results they have obtained to one another. In other DAI text using a layered approach has also been considered. In [38] for instance, the author proposes breaking the agents into two categories: supervisory or high-level, and local or low-level agents. If the relationship between these two levels are of a dependence ($\leftarrow$) nature as described before, the order of the hyper-structure resulting from this layered approach would be $2 + r$, where $r$ is the structural order of the agents.
The problem here would be to motivate the agents into automatically acquiring hierarchical roles in the population. Although these works are usually either theoretical, or not based on machine learning, they can serve as a good basis for practical machine learning SOSAGEs. The mere fact that this structural approach, be it hand coded, has improved the performance of such systems is encouraging.

In the field of Neural Networks, some promising work has been done recently on self-organizing networks. One hot topic deals with developing ways to specify the topology of neurons and their links in multi-layer neural networks. This process was largely done by trial and error methods resulting in sub-optimal networks with too many units or links, or taking too much time to train, or not converging at all. For example in GNARL (Generalized Acquisition of Recurrent Links) connections, weights and the number of hidden layers of a neural network are defined using evolutionary programming techniques [1]. This method, although self-organizing to a certain extent, results in a hyper-structure of the second order (the nodes serving as the first order structure and the network as a whole being its second order). These along with other similar methods ([10] and [16]) use genetic methods in which each chromosome encodes a complete network. In other words, the whole Hyper-structure is defined not allowing for the further break down of the problem past the network level.

In SANE (Symbiotic Adaptive neuro-Evolution) individual neurons are evolved to form complete neural networks [29]. Each neuron or group of neurons is treated as a single structure being evaluated separately. The resulting sub-networks are then grouped together and evaluated in complete neural networks. Thus the system is actually of an order higher than $2^3$. Figure 2 shows a hyper-structure visualization of a similar system similar to SANE. Because no single neuron can perform well alone, the population remains diverse and the genetic algorithm can search in many different areas of the solution space concurrently. SANE can thus find solutions faster, and to harder problems than other standard neuro-evolution systems.

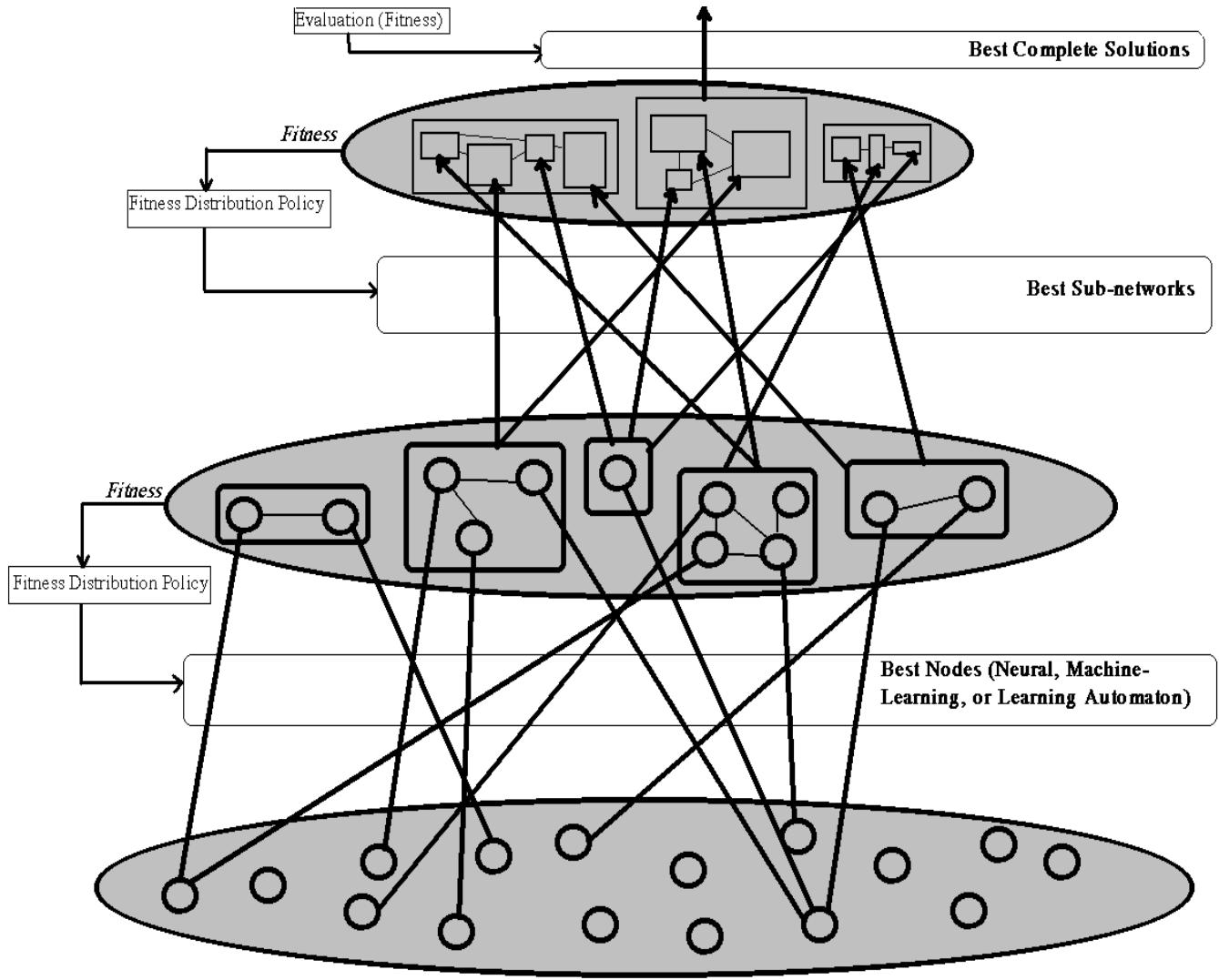

**Fig.2.** A 3$^{rd}$ order hyper-structure representation of a system similar to SANE. Only the fitness returned from the highest level is concrete and there must be certain policies to distribute this fitness between individual agents from the lower level.

---

[3] In the case of the references no hyper-structure of order higher than 4 has been implemented, but the idea can easily be upgraded to dynamically scale up to higher orders by increasing the evolutionary optimization stages

SANE can solve easy tasks in just a few generations, while in tests that require high precision, its progress often stalls and is exceeded by standard network-level evolution. In [30] a hierarchical version of SANE is introduced that uses two levels of evolution and overcomes the problem noted above. In this method, two loops of genetic evolution are used, increasing the order of the hyper-structure. Some improvements on the method used to evaluate the sub-networks have also been made.

The SANE approach could be thought of as evolving *neural sub-networks* that best fit a multitude of complete neural networks. If each neuron is taken as a structure of order 1, the sub-networks are examples of second order structures and the neural networks are hyper-structures of order 3. Note that we could have complete neural networks evolving along side sub-networks and neurons. Therefore this increase in structural order is in fact happening automatically and without outside intervention. If, unlike the references, the number of hidden layers were not limited to one, and sub networks were allowed to organize into larger sub-networks, the order of solution hyper-structures could then be expected to be limitless (or limited to the population limit).

In general, Reinforcement Learning (RL) work has concentrated on problems with a single goal. As the complexity of the problem scale up, both the size of the state-space and the complexity of the reward function increase. We will clearly be interested in methods of breaking problems up into sub-problems, which can work with smaller state-spaces and simpler reward functions, and then having some method of combining the sub-problems to solve the main task [19]. Most of the work in RL either designs the decomposition by hand [28], or deals with problems where the sub-tasks have termination conditions and combine sequentially to solve the main problem [36] [39].

Behavior based reinforcement learning methods (e.g., Q-Learning) have been used successfully for robot design. Increasing robot complexity makes this method of design difficult. LISP like programs, neural networks, and classifier systems have all been evolved as robot controllers before (see [12] for a survey of these methods). The main drawbacks of these methods have been sited as their inability to scale-up to more difficult behavior. In [24] an evolutionary method is suggested to provide design automation for building behavioral modules. In this paper the behavior primitives and arbitrators are evolved instead of being hand-coded. A similar method based on [40]'s work on modular Q-Learning uses adaptive methods to automate the modulation of the Q-Learning Architecture [22]. The Action Selection problem introduced by [19] essentially concerns subtasks acting in parallel, and interrupting each other rather than running to completion. Typically, each subtask can only ever be partially satisfied [26]. In this method different parts of the society of agents modify their behavior based on whether or not they are succeeding in getting the top-level action selector to execute their actions. The learning of the subtasks is again left to an RL algorithm so that there is two levels of reinforcement learning being conducted simultaneously. [25] also proposes a hierarchical Q-Learning method in which a Q-Learner learns to choose from actions proposed by lower level Q-learners. In this and the previous method ([19]) the user designs the sub-problems.

In the cases mentioned here, hand coded hierarchies (which are subsets of hyper-structures according to [2]) are given the power of automatically changing their configurations. Again the fact that higher order hyper-structures (in this case no more than 2) are shown to be more effective is encouraging. Leaving the system to define the behavior automatically and dynamically also opens the chance for the hyper-structure to decide not to break-up. This would occur if only one behavior, which is that of solving the main problem, is chosen.

## 6. Further research

Organic life and higher-order organisms viewed as evolutionary products are an example of known structures that may naturally be considered as hyper-structures. Therefore it is only natural to try to use Artificial Evolutionary methods to create SOSAGEs. It has been shown [31] that the combination of evolutionary methods with machine learning methods results in much faster and more efficient solutions, especially in agent-based systems. The method used by the agents in their lifetime need not necessarily be machine learning. Simple rule-based or classifier systems could also be used, but the non-parametric machine learning agents are much more flexible to dynamic environmental change [27]. It must be noted that although the problem itself may be of a static nature, the group approach implied by the SOSAGE method will be constantly changing as different agents improve their approach to the problem. Other useful agent architectures could be Behavior Based Q-Learning, Artificial Neural Networks, and Memory Based machine learning agents [17].

The Evolutionary method, although very effective, is not a pre-requisite of SOSAGE systems. Populations of agents that have flexible communication and collaboration capabilities can also be used to develop SOSAGE systems. Here are certain areas worth considering [23]:

- Game Theory
- Coalition Formation [6] and Cooperation Structures [8]
- Classical Mechanics
- Operational Research
- Social Sciences [13] [9] [33]

A desirable feature that enhances the generality and flexibility of SOSAGEs is the agents being as problem independent as possible. The ultimate of such a system would be for the user only to define the problem and leave it up to the agents to solve it. This has been described before as the TEEZ-HOOSH approach [18].

## 7. Conclusions

This paper suggests a hyper-structure approach to modeling and solving problems and creating agents. This approach will attempt to break down the original solution hyper-structure until the smallest units are capable, efficiently, of maximizing their reward function. Single agents, until now thought of as units of decision-making, can be seen as self-structured populations of hierarchically arranged sub-agents or SOSAGEs.

Although the idea in itself is new, we have shown that work in this area has already started and preliminary results seem encouraging.

The main purpose of this paper is to give an introduction to our ideas on engineering problem solving methods that incorporate emergence and hyper-structures. This idea is at an initial stage and needs further elaboration and experimentation.